\documentclass[pdflatex,sn-mathphys-num]{sn-jnl}% Math and Physical Sciences Numbered Reference Style
\usepackage{graphicx}%
\usepackage{multirow}%
\usepackage{amsmath,amssymb,amsfonts}%
\usepackage{amsthm}%
\usepackage{mathrsfs}%
\usepackage[title]{appendix}%
\usepackage{xcolor}%
\usepackage{textcomp}%
\usepackage{manyfoot}%
\usepackage{booktabs}%
\usepackage{algorithm}%
\usepackage{algorithmicx}%
\usepackage{algpseudocode}%
\usepackage{listings}%
\usepackage{gensymb}

\usepackage{todonotes}

\usepackage[nolist,nohyperlinks]{acronym}
\acrodef{ar}[AR]{Augmented Reality}
\acrodef{auc}[AUC]{Area Under the Curve}
\acrodef{mis}[MIS]{minimally invasive surgery}
\acrodef{ramis}[RAMIS]{robotic-assisted minimally invasive surgery}
\acrodef{hsi}[HSI]{hyperspectral imaging}
\acrodef{sfm}[SfM]{structure-from-motion}
\acrodef{slam}[SLAM]{Simultaneous Localisation And Mapping}
\acrodef{2d}[2D]{two-dimensional}
\acrodef{3d}[3D]{three-dimensional}
\acrodef{fps}[fps]{frames-per-second}
\acrodef{hsspn}[HSSPN]{HyperSpectral SuperPoint Network}
\acrodef{sp}[SP]{SuperPoint}
\acrodef{ha}[HA]{homographic adaptation}
\acrodef{hha}[HHA]{hyperspectral homographic adaptation}
\acrodef{sift}[SIFT]{Scale-Invariant Feature Transform}
\acrodef{ransac}[RANSAC]{random sample consensus}
\acrodef{lift}[LIFT]{Learned Invariant Feature Transform}
\acrodef{cnn}[CNN]{convolutional neural network}
\acrodef{dl}[DL]{deep learning}
\acrodef{fast}[FAST]{Features from Accelerated Segment Test}
\acrodef{surf}[SURF]{Speeded-Up Robust Features}
\acrodef{dog}[DoG]{Difference-of-Gaussians}
\acrodef{orb}[ORB]{Oriented FAST and Rotated BRIEF}
\acrodef{alike}[ALIKE]{Accurate and Lightweight Keypoint Detection and Descriptor Extraction}
\acrodef{aliked}[ALIKED]{A LIghter Keypoint and Descriptor Extraction with Deformable transformation}
\acrodef{nms}[NMS]{non-maximum suppression}
\acrodef{dkd}[DKD]{Differentiable Keypoint Detection}
\acrodef{nn}[NN]{Nearest Neighbour}
\acrodef{maa}[mAA]{Mean Average Accuracy}
\acrodef{mnn}[MNN]{Mutual Nearest Neighbour}
\acrodef{dsm}[DSM]{Dual Softmax Matcher}
\acrodef{gnn}[GNN]{graph neural network}
\acrodef{roma}[RoMa]{Robust Dense Feature Matching}
\acrodef{dexined}[DexiNed]{Dense Extreme Inception Network for Edge Detection}
\acrodef{dml}[DML]{Dual-Modality Laparoscope}
\acrodef{ms}[MS]{Matching Score}
\acrodef{mma}[MMA]{Mean Matching Accuracy}
\acrodef{mha}[MHA]{Mean Homography Accuracy}
\acrodef{rgb}[RGB]{red-green-blue}
\acrodef{hs-alike}[HS-ALIKE]{HyperSpectral-ALIKE}
\acrodef{prgb}[pRGB]{pseudo-red-green-blue}
\acrodef{nerf}[NeRF]{neural radiance field}
\acrodef{ss-sift}[SS-SIFT]{spectral-spatial-SIFT}
\acrodef{umsgc-sift}[UMSGC-SIFT]{Unified Model of Spectral Value and Gradient Change SIFT}
\theoremstyle{thmstyleone}%
\theoremstyle{thmstyletwo}%

\theoremstyle{thmstylethree}%

\begin{document}

\title[HyKey]{HyKey: Hyperspectral Keypoint Detection and Matching in Minimally Invasive Surgery}


\author*[1,2]{\fnm{Alexander} \sur{Saikia}}\email{alexander.saikia.21@ucl.ac.uk}

\author[1]{\fnm{Chiara} \sur{Di Vece}}

\author[1]{\fnm{Zhehua} \sur{Mao}}

\author[1]{\fnm{Sierra} \sur{Bonilla}}

\author[1]{\fnm{Chloe} \sur{He}}

\author[1,2]{\fnm{Joao} \sur{Ramalhinho}}

\author[1,2]{\fnm{Tobias} \sur{Czempiel}}

\author[1]{\fnm{Sophia} \sur{Bano}}

\author[1]{\fnm{Danail} \sur{Stoyanov}}

\affil[1]{\orgdiv{UCL Hawkes Institute, Dept of Medical Physics and Biomedical Engineering and Dept of Computer Science}, \orgname{University College London}, \orgaddress{\city{London}, \postcode{WC1E 6BT}, \country{United Kingdom}}}

\affil[2]{\orgdiv{EnAcuity Ltd.}, \orgaddress{\city{London}, \postcode{EC2A 4NE}, \country{United Kingdom}}}


% \abstract{The abstract serves both as a general introduction to the topic and as a brief, non-technical summary of the main results and their implications. Authors are advised to check the author instructions for the journal they are submitting to for word limits and if structural elements like subheadings, citations, or equations are permitted.}

%%================================%%
%% Sample for structured abstract %%
%%================================%%

\abstract{
\textbf{Purpose:} 3D reconstruction in minimally invasive surgery (MIS) enables enhanced surgical guidance through improved visualisation, tool tracking, and augmented reality. However, traditional RGB-based keypoint detection and matching pipelines struggle with surgical challenges, such as poor texture and complex illumination.
% , including poor texture and complex illumination. 
We investigate whether using snapshot hyperspectral imaging (HSI) can provide improved results on keypoint detection and matching surgical scenes.

\textbf{Methods:} We developed HyKey, a HYperspectral KEYpoint detection and description model made up of a hybrid 3D-2D convolutional neural network that jointly extracts spatial-spectral features from HSI. The model was trained using synthetic homographic augmentation and epipolar geometry constraints on a robotically-acquired dual-camera RGB-HSI laparoscopic dataset of ex-vivo organs with calibrated camera poses. We benchmarked performance against established RGB-based methods, including SuperPoint and ALIKE.

\textbf{Results:} Our HSI-based model outperformed RGB baselines on 
% both pseudo-RGB (extracted from HSI) and 
registered RGB frames, achieving 96.62\% mean matching accuracy and 67.18\% mean average accuracy at 10$\degree$ on pose estimation, demonstrating consistent improvements across multiple evaluation metrics.

\textbf{Conclusion:} Integrating spectral information from an HSI cube offers a promising approach for robust monocular 3D reconstruction in MIS, addressing limitations of texture-poor surgical environments through enhanced spectral-spatial feature discrimination. Our model and dataset are available at \url{https://github.com/alexsaikia/HyKey-Hyperspectral-Keypoint-Detection}
}

\keywords{Hyperspectral imaging, Keypoint detection and description, 3D reconstruction, Minimally invasive surgery}
 


\maketitle

\section{Introduction}
\label{sec:intro}

\Ac{hsi} is an emerging modality with strong potential for intraoperative use~\cite{lu2014medical,clancy2020surgical,ali2025surgical}. By combining imaging and spectroscopy, \ac{hsi} enables the extraction of both spatial and biochemical information, such as tissue composition, haemoglobin saturation, and morphology, supporting identification of pathological changes, tissue segmentation, and guidance during interventions where subtle spectral differences are invisible to both human vision and standard \ac{rgb} cameras. Traditional \ac{rgb} endoscopic imaging does not provide sufficient biochemical or morphological information to support these advanced surgical tasks. Moreover, \ac{hsi} alone is not sufficient in the operating room, as surgeons need both geometry and physiology. If we can reconstruct the operative field in 3D and attach spectral information per surface patch, we can visualise perfusion heterogeneity around anastomoses, support margin assessment, and stabilise \ac{ar} overlays, providing capabilities beyond standard \ac{rgb} endoscopy.  

While 3D reconstruction techniques like \ac{sfm}~\cite{ullman1979interpretation,tomasi1992shape} and \ac{slam}~\cite{durrant2006simultaneous} have been combined with \ac{dl} to improve accuracy and robustness in \ac{rgb} computer vision tasks, adapting these methods to \ac{mis} challenges, such as tissue deformation, specular highlights, poor texture, occlusions, and varying illumination, remains difficult. As a result, keypoints detection becomes unstable and descriptors drift, which degrades matching and downstream \ac{sfm}/\ac{slam}. Reliable ground truth for evaluation, such as absolute camera poses or 3D point clouds, is also challenging to obtain. Combining 3D reconstruction with spectral data offers the possibility of enriched surgical maps that visualise anatomy in 3D while providing biochemical context to guide surgical decisions.

Keypoint detection and description are a fundamental step in 3D reconstruction, with classical hand-crafted methods such as \ac{sift}~\cite{lowe2004distinctive}, \ac{surf}~\cite{bay2008speeded}, and \ac{orb}~\cite{rublee2011orb} forming the basis for matching and \ac{sfm} in traditional \ac{rgb} imaging. While these methods offer scale invariance and efficiency, they remain sensitive to viewpoint changes, non-rigid deformation, and challenging photometric conditions common in surgical scenes, leading to unstable keypoints and degraded matching. With advances in \ac{dl}, more sophisticated detectors have emerged, including \ac{sp}~\cite{detone2018superpoint}, \ac{alike}~\cite{zhao2022alike}, and \ac{aliked}~\cite{zhao2023aliked}, which have improved spatial localisation by learning more robust features from large annotated or synthetic datasets. However, these \ac{rgb}-based detectors are fundamentally built on 2D convolutions and are limited in their ability to process the full spectral content of an \ac{hsi} cube. The only prior work extending learning-based detection and description to hyperspectral data, \ac{hsspn}~\cite{ma2022deep}, remains restricted to 2D convolutional architectures, treating each spectral band as an independent channel. While 2D CNNs mix channels through learned weights, stacking bands as channels treats wavelengths like a set of generic channels rather than an explicit measurement axis. Since an HSI cube is naturally represented as a 3D volume, prior work has argued that 3D filtering provides a simple way to jointly extract spectral-spatial features jointly and hybrid 3D-2D designs use early 3D layers to form joint representations before switching to 2D processing for efficiency~\cite{roy2019hybridsn,noshiri2023comprehensive}.

To overcome these limitations, we propose HyKey, a spectral-spatial detector and descriptor for \ac{hsi} in \ac{mis} with a 3D-2D backbone and differentiable keypoint detection, trained via homography and epipolar supervision on a dual-modality \ac{rgb}-\ac{hsi} dataset (robotic \textit{ex-vivo} and \textit{in-vivo} videos). Our main contributions are as follows:

\begin{itemize}
    \item \textbf{Spectral-spatial features for \ac{hsi}:} We design a compact 3D-2D CNN that learns inter-band spectral-spatial cues and outputs a score map, keypoints and dense descriptors with a differentiable keypoint detection head for end-to-end training. 
    \item \textbf{Geometry-aware supervision:} Our training couples synthetic homography objectives with an epipolar regulariser to minimise Sampson distances without loss of differentiability, leading to improved relative pose recovery. 
    \item \textbf{Dual-modality laproscopic dataset}: We acquire two dual-modality (\ac{hsi} + registered \ac{rgb}) datasets in robotic \textit{ex-vivo} laparoscopic organ imagery (released with this work) and in-vivo colorectal procedures, enabling benchmarking of \ac{hsi} matching vs \ac{rgb} baselines.
    \item \textbf{Comprehensive benchmarking:} We report a detailed evaluation showing that HyKey improves homography metrics and relative pose over \ac{rgb} and \ac{hsi} baselines.
\end{itemize}

\section{Methodology}\label{sec:Methods}

\subsection{Dual-Modality Dataset Description}
\label{subsec:datasets}

\begin{figure}[t]
    \centering
    \includegraphics[width=\linewidth]{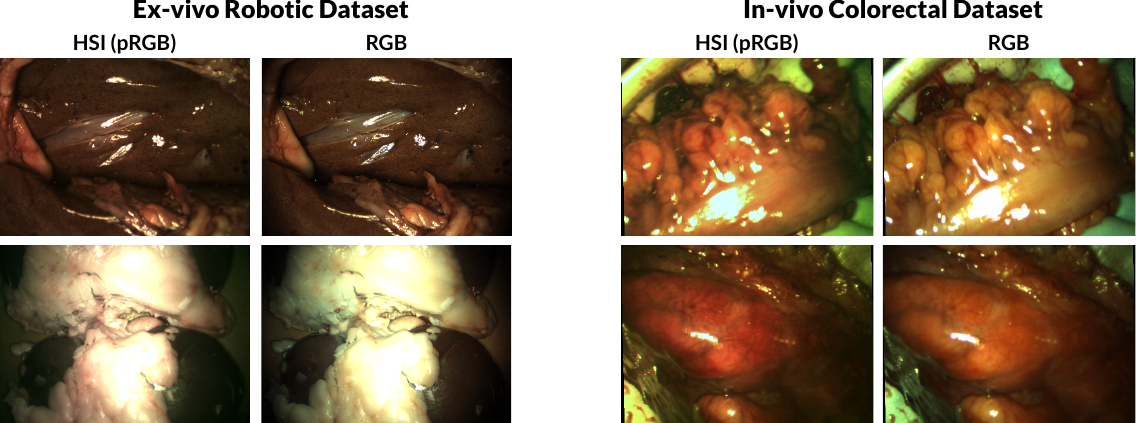}
    \caption{Example frames from the proposed dual-modality datasets. Each includes co-registered hyperspectral (visualised as pRGB) and \ac{rgb} images. The close spatial alignment between modalities ensures reliable evaluation between \ac{rgb} baselines and \ac{hsi} models}.
    \label{fig:dataset}
\end{figure}

To train, validate, and benchmark our model and baselines, we utilise two in-house datasets acquired using a custom \ac{dml} as illustrated in Figure~\ref{fig:dataset}.\\

\paragraph{Dual-Modality Laparoscope} 
We acquired two datasets with the same custom-built \ac{dml}, equipped with both an \ac{hsi} and an \ac{rgb} camera. The \ac{hsi} camera (MQ022HG-IM-SM4x4, XIMEA GmbH, Muenster, Germany) is a $4~\times~4$ mosaic snapshot camera capturing 16 spectral bands from 460 to 600nm and a resolution after demosaicing to $272~\times~512~\times~16$.
The \ac{rgb} camera is a FLIR Blackfly S USB3 (BFS-U3-50S5C-C, Teledyne FLIR, Wilsonville, Oregon, USA), with a raw resolution of $2448 ~\times~ 2048$.
Both cameras are attached to a beamsplitter that is custom-mounted onto a HOPKINS\textsuperscript{\textregistered} Telescope 26003 AGA (Karl Storz SE \& Co. KG, Tuttlingen, Germany), therefore aligning their principal axes. Our in-house acquisition software captures a pair of corresponding \ac{rgb} and \ac{hsi} frames at each time point. 
Due to the different aspect ratios and camera orientations, we register the \ac{rgb} to the \ac{hsi} cube using keypoints from a Charuco calibration dataset. This enables fair comparison between both modalities.

\paragraph{In-house Ex-vivo Robotic Dataset}
Our in-house robotic dataset was collected using a camera-agnostic robotic arm platform~\cite{saikia2025robotic} using a 7 degrees-of-freedom manipulator to position the laparoscope around \textit{ex-vivo} ovine organs. Eye-in-hand calibration was performed using a Charuco board as per the paper.
Following~\cite{saikia2025robotic}, we collected synchronised \ac{rgb}-\ac{hsi} video sequences under the same multiple lighting and trajectory scenarios. The dataset includes 10 ovine kidneys, with some paired up to form a total of 7 organ sets, and 3 ovine livers, and includes multiple recordings of each organ under different acquisition scenarios.

All acquisitions were performed using a 0\degree~\,30\,cm 10\,mm diameter HOPKINS laparoscope.
The different lighting scenarios involved an open surgical light source or the laparoscopic lightsource. The different trajectory scenarios involved both unconstrained, externally controlled hemispherical sweeps and constrained movements mimicking the remote centre of motion about a trocar port.

In total, the dataset consists of 52 videos, for a total of 41,449 \ac{rgb}-\ac{hsi} frame pairs. For evaluation, videos of one kidney and one liver were held out as an unseen test set (9,150 frames), and a different kidney–liver pair was reserved as a validation control (5,544 frames).
While the \textit{ex-vivo} setting differs from \textit{in-vivo} \ac{mis} conditions, the dataset effectively captures the spatial–spectral characteristics relevant for keypoint learning, making it well-suited for training and evaluation. We are releasing this dataset alongside this work.

% \noindent \textbf{In-house In-vivo Colorectal Dataset:} 
\paragraph{In-house In-vivo Colorectal Dataset} 
To address the domain gap associated with the robotic \textit{ex-vivo} dataset, we additionally collected an \textit{in-vivo} dataset from 12 patients consented for laparoscopic colorectal cancer resection. 
The study received ethical approval from the Yorkshire \& The Humber – Bradford Leeds Research Ethics Committee (Reference: 24/YH/0088; IRAS project ID: 331933) and was sponsored by Imperial College London. 
Hyperspectral and \ac{rgb} images were acquired from multiple views of the tumour region, both \textit{in-vivo} and \textit{ex-vivo} after resection. We removed two patients' videos for the test set (11,912 frames), a single patient for the validation set (8,711 frames) with the training set containing the other nine patients (61,995 frames).
Due to patient confidentiality, this dataset is not available for public release.

\subsection{Proposed HyKey model}
\label{subsec:hykey-model}

Given an \ac{hsi} cube $I\in\mathbb{R}^{C,H,W}$ (with $C=16$ bands in our system), HyKey aims at (i) detecting a set of repeatable keypoints and (ii) computing robust local descriptors so that correspondences between views can be established reliably. These correspondences can then feed standard geometric estimators (e.g., homography, essential matrix) to enable 3D reconstruction and camera tracking in \ac{mis} scenes. %Figure~\ref{fig:architecture} summarises our architecture. 

\begin{figure}[t]
    \centering
    \includegraphics[width=1\linewidth]{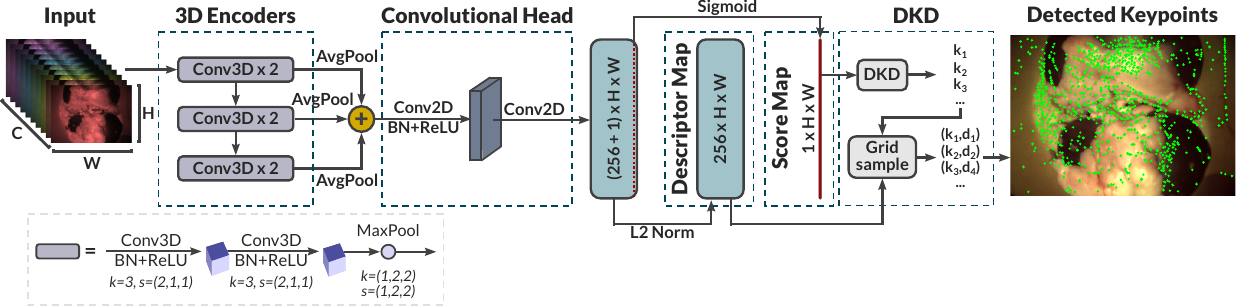}
    \caption{Architecture of the proposed HyKey network for keypoint detection and description in \ac{hsi}. The network uses three sequential 3D convolutional encoder blocks to extract spatial-spectral features, followed by feature aggregation and upsampling. A 2D convolutional head produces an L2-normalised descriptor map and a keypoint score map, which feeds a DKD module for end-to-end trainable keypoint extraction.}
    \label{fig:architecture}
\end{figure}

HyKey adopts a hybrid 3D-2D CNN tailored to \ac{hsi} (Figure~\ref{fig:architecture}). 3D convolutions capture spectral–spatial correlations early, while a lightweight 2D head produces a keypoint score map and dense descriptors at full image resolution. We prefer this design over commonly reported transformer-heavy models for other \ac{hsi} tasks~\cite{kumar2024deep,tejasree2024extensive} because surgical \ac{hsi} training data are scarce and real-time constraints make large transformers impractical in the operating room. In addition, to the authors' knowledge, there are currently no transformer-based models focusing on hyperspectral keypoint detection and matching.

\subsubsection*{Architecture Details}

% \noindent \textbf{3D Spectral-Spatial Encoder:} 
\paragraph{3D Spectral-Spatial Encoder}
The \ac{hsi} cube $I$ passes through three 3D convolution blocks, with each block containing two 3D convolutional layers with kernel size $3~\times~3~\times~3$ and stride $(2,1,1)$, thus halving the spectral bands at each convolution. Each block outputs a different number of channels $c_i$ ($c_1=32$, $c_2=64$, $c_3=128$), with both convolutional layers within each block producing the same number of output channels. After the two convolutions, a spatial max pooling operation (kernel $1,2,2$) is applied. This produces an output shape of $[c_i,C_{in}/4,H_{in}/2,W_{in}/2]$.

% \noindent \textbf{Feature Aggregation:} 
\paragraph{Feature Aggregation}
The outputs from the three encoder blocks are aggregated using an average pooling across the spectral dimensions, collapsing them to a single channel. After squeezing this singleton dimension, each of the three feature blocks has shape $[c_i,H/2^i,W/2^i]$. These three feature maps are then upsampled to the original spatial resolution $(H_{\text{orig}}, W_{\text{orig}})$ via bilinear interpolation and concatenated along the channel dimensions, producing an aggregated feature block of $F \in \mathbb{R}^{(c_1+c_2+c_3)~\times~ H_{\text{orig}}~\times~ W_{\text{orig}}}$.

% \noindent \textbf{2D Convolutional Head:} 
\paragraph{2D Convolutional Head}
We apply two $3~\times~3$ convolutions to $F$ to obtain detection scores and descriptors. The first convolution produces $D$ channels and is followed by batch normalisation and ReLU. The second convolution produces $D{+}1$ channels, which are split into a score map $S$ (sigmoid activation) and a $D$-channel descriptor map; the descriptor vectors are $L2$-normalised per pixel.

% \noindent \textbf{Differentiable Keypoint Detection:} 
\paragraph{Differentiable Keypoint Detection}
The score map $S$ is fed to the \ac{dkd} module as detailed in~\cite{zhao2022alike,zhao2023aliked}. Unlike conventional approaches that apply \ac{nms} directly to the score map, which breaks differentiability and prevents end-to-end training, the \ac{dkd} module enables differentiable keypoint extraction through soft selection within a local window. This preserves gradient flow throughout the network, allowing joint optimisation of the encoder, decoder, and keypoint detection stages. Descriptors are extracted by grid sampling the descriptor map at the keypoint spatial location.

\subsection{Model Optimisation}\label{sec:loss_functions}
\begin{figure}[t]
    \centering
    \includegraphics[width=0.92\linewidth]{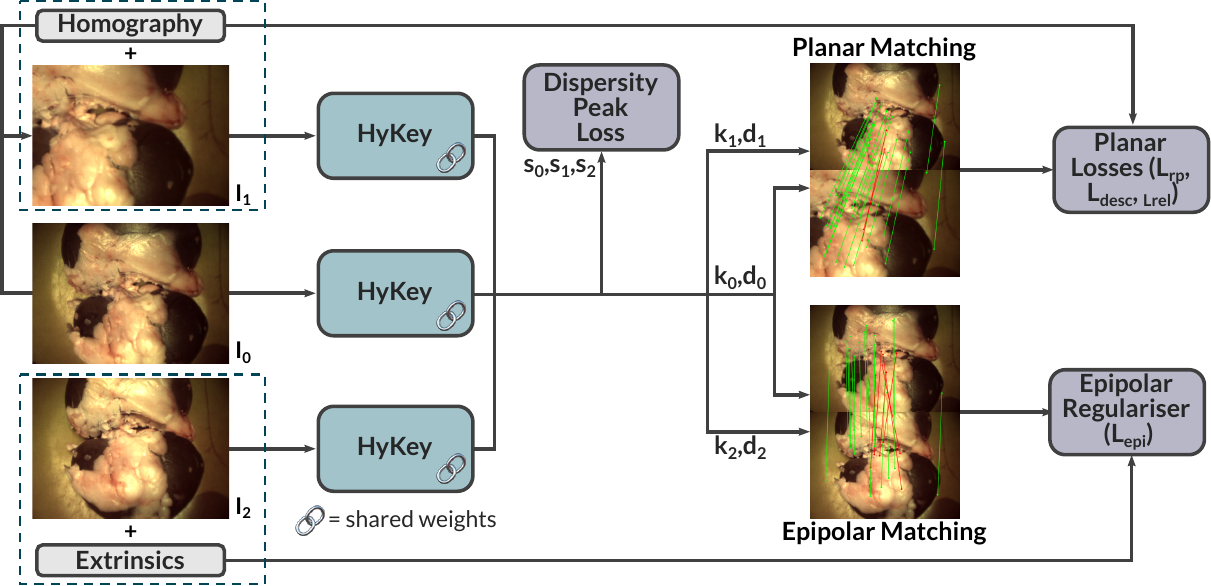}
    \caption{Overview of the HyKey training pipeline. An input frame $I_0$ is paired with a homographically warped version $I_1$ for synthetic supervision. For the robotic dataset, an additional sequential frame $I_2$ from the same video provides a real second view. All inputs are processed by HyKey individually to extract $s_i,k_i,d_i$, score maps, keypoints, and descriptors, which are jointly optimised using planar and epipolar matching constraints.}
    \label{fig:trainstep}
\end{figure}

Each training sample consists of three images: a base view \(I_0\), a planar homographic view \(I_1~=~W_{\mathbf{H}_{01}}(I_0)\), and an epipolar view \(I_2\) (for robotic data only) with known camera intrinsics $K$ and extrinsics \((\mathbf{R}_{02}, \mathbf{t}_{02})\). For each image, the network predicts keypoints \(\mathcal{K}_i\), descriptors \(\mathcal{D}_i\), and a score map \(\mathbf{S}_i\). Descriptor correspondences are established through \ac{mnn} matching to maintain matcher-agnostic consistency. One-to-one matches are enforced by retaining only mutually consistent correspondences, where keypoint $i$ from one image matches keypoint $j$ from another image if and only if $j$ matches back to $i$; keypoints without such mutual matches are excluded. HyKey is only trained on \ac{hsi} data under the mixed planar-epipolar supervision regime described below. %in Section~\ref{subsec:datasets}.
An overview of the training pipeline is presented in Figure~\ref{fig:trainstep}. 
We combine a planar supervision term from \((I_0,I_1)\) with a geometry-aware term from \((I_0,I_2)\) to get the total loss as:
\begin{equation}
\mathcal{L}_{\text{total}}~=~\lambda_{\text{pk}}\,\mathcal{L}_{\text{pk}}
+ \lambda_{\text{rp}}\,\mathcal{L}_{\text{rp}}
+ \lambda_{\text{rel}}\,\mathcal{L}_{\text{rel}}
+ \lambda_{\text{desc}}\,\mathcal{L}_{\text{desc}}
+ \lambda_{\text{epi}}\,\mathcal{L}_{\text{epi}}.
\end{equation}
This joint formulation balances appearance, localisation, and geometric consistency across planar and epipolar pairs. Each loss term is detailed below and is weighted as follows: $\lambda_{\text{pk}}~=~0.5$, $\lambda_{\text{rp}}~=~1$, $\lambda_{\text{rel}}~=~1$, $\lambda_{\text{desc}}~=~5$, and $\lambda_{\text{epi}}~=~0.25$, with the planar loss weights inspired by~\cite{zhao2023aliked} and the epipolar loss set empirically.

\textbf{Dispersity peak loss \(\mathcal{L}_{\text{pk}}\)} is applied independently to each image's score map \(\mathbf{S}_i\) (\(i~\in~\{0,1\}\)), encouraging sharp and localised peaks corresponding to keypoints. It penalises diffuse activations within a \(5~\times~5\) neighbourhood, using a sigmoid-based weighting of the keypoint confidence centred at a score threshold \(t_{\text{sc}}\). This term promotes spatially distinct and reliable detections prior to any geometric supervision.

\textbf{Planar homography losses (\(\mathcal{L}_{\text{rp}}, \mathcal{L}_{\text{desc}},\mathcal{L}_{\text{rel}}\)):}
For the planar image pair \((I_0, I_1)\), we adopt an ALIKE-style self-supervised framework. Keypoints from \(I_0\) are projected to \(I_1\) via the known homography \(\mathbf{H}_{01}\) to establish pseudo-ground-truth correspondences, while random locations are additionally sampled to promote descriptor diversity. Three geometric losses are applied: (i) the reprojection localisation loss \(\mathcal{L}_{\text{rp}}\) penalises geometric inconsistency between projected and detected keypoints using a Huber loss on the reprojection error, weighted by the product of sigmoid weights from both views; (ii) the sparse score map reliability loss \(\mathcal{L}_{\text{rel}}\) encourages confidence consistency between matching keypoints across score maps and descriptor similarities; and (iii) the descriptor reprojection loss \(\mathcal{L}_{\text{desc}}\) aligns descriptors across homographic pairs by optimizing a cosine-similarity matrix via cross-entropy with symmetric supervision.

\textbf{Geometry-aware epipolar loss \(\mathcal{L}_{\text{epi}}\):} For the non-planar image pair \((I_0, I_2)\), geometric consistency is enforced through the fundamental matrix \(\mathbf{F}_{02}~=~\mathbf{K}_{2}^{-T}[\mathbf{t}_{02}]_\times \mathbf{R}_{02}\mathbf{K}_{0}^{-1}\), which constrains corresponding keypoints to lie on epipolar lines. The epipolar regulariser \(\mathcal{L}_{\text{epi}}\) computes soft correspondence probabilities from descriptor similarities and minimises the expected Sampson error over candidate correspondences using a Huber loss symmetrically in both directions. This encourages robust descriptor learning across non-planar, tissue-like surfaces and complements the planar geometric supervision.

\section{Experimental Setup}

For this work, we train two models: HyKey-noPE (no positional epipolar regulariser enabled) and HyKey. For HyKey, it was activated after the first 5 epochs to allow stable descriptor learning. During training, a maximum of 400 detected keypoints with the highest scores and 400 random keypoints were used. The DKD and planar loss hyperparameters followed the configuration from~\cite{zhao2023aliked}.

The models were trained using the Adam optimiser with a learning rate of $3~\times~10^{-4}$, linearly warmed up over the first 500 steps.
A batch size of 6 was used, and to account for varying dataset sizes, training was capped at 10,000 frames per epoch with half of each batch coming from each dataset.
All experiments were conducted on a workstation equipped with an AMD Ryzen Threadripper PRO 5975WX CPU and a single NVIDIA RTX A6000 GPU, requiring approximately 43 hours for one training run.

% \noindent \textbf{Evaluation metrics:} 
\subsection{Evaluation Metrics}
Following previous work~\cite{detone2018superpoint,balntas2017hpatches,zhao2022alike,zhao2023aliked} we utilise the following planar metrics:
\begin{itemize}
\item \textbf{Repeatability}: proportion of keypoints that have the possibility to be correctly re-detected in the corresponding image within a pixel threshold $\tau$.
\item \textbf{\ac{ms}}: proportion of keypoints successfully matched between image pairs under a pixel threshold $\tau$.
\item \textbf{\ac{mma}}: proportion of matches with reprojection error below $\tau$, indicating geometric matching accuracy.
\item \textbf{\ac{mha}}: proportion of image corners correctly aligned within $\tau$ after applying the homography estimated from the predicted matches.
\end{itemize}

The repeatability and \ac{ms} metrics are proportional to the total number of keypoints that would be co-visible in each image.
For relative pose, we evaluate \textbf{\ac{maa}}, representing the proportion of predicted poses with angular error below a threshold.\\

% \noindent \textbf{Model comparison and evaluation settings:} 
\subsection{Model Comparison and Evaluation Settings}
We evaluate our HyKey model against both trained/finetuned \ac{hsi} models (HSSPN, HS-ALIKE) and \ac{rgb} baselines (\ac{sift}, \ac{sp}, \ac{alike}) on the \textit{in-vivo} colorectal and \textit{ex-vivo} robotic test sets. For parity across modalities, we register \ac{rgb} frames to the \ac{hsi} reference, then crop both to remove invalid regions. The registration direction was chosen so as to not introduce a confounding advantage towards the higher resolution \ac{rgb} camera.For evaluation, the model and baselines were set to return a maximum number of 1024 keypoints.
We evaluate without test-time augmentation and use \ac{mnn} matching to ensure fair comparison, avoiding additional supervision from learned neural matchers.
We report two complementary settings:
\begin{enumerate}
    \item Synthetic homography tests on both test sets with Rep, \ac{ms}, \ac{mma}, \ac{mha} at thresholds of $\tau~=~[1,3,5,10,20]$ and evaluate the \ac{auc}.
    \item Real video pairs (\textit{ex-vivo} robotic test set with ground truth poses) with mAA@{5$\degree$,10$\degree$,20$\degree$} using robust estimator USAC-MAGSAC~\cite{barath2020magsac++} (threshold~=~1 normalised px, confidence~=~0.99999) to obtain the fundamental matrix.
\end{enumerate}

\section{Results and Discussion}\label{sec:results}
\subsection{Loss Function Ablation}
\label{subsec:loss-ablation}

We ablate the loss components in Eq.~(1) to quantify their effects on planar localisation/matching and non-planar geometric consistency. We evaluate (i) Repeatability-AUC and \ac{ms}-AUC under synthetic homography warps on both datasets, and (ii) relative pose accuracy (mAA@10$\degree$) on real video pairs from the \textit{ex-vivo} robotic dataset.

\begin{table}[t]
\centering
\caption{Loss ablation on the test sets. Rep/MS are AUC (\%) under synthetic homographies; mAA@10$\degree$ is relative pose accuracy (\%) on real video pairs solely from the \textit{ex-vivo} robotic dataset (best \textbf{bold}, second \underline{underlined}). $\checkmark$~indicates the corresponding loss term in Eq.~(1) is enabled.}
\label{tab:loss-ablation}

\begin{tabular*}{\linewidth}{@{\extracolsep{\fill}}lccccc|ccc@{}}
\toprule
\multirow{2}{*}{\textbf{Model}} & \multicolumn{5}{c|}{\textbf{Loss terms enabled (Eq.~(1))}} & \multicolumn{3}{c}{\textbf{Metrics}} \\
\cmidrule(lr){2-6}\cmidrule(lr){7-9}
& $\mathcal{L}_{\text{pk}}$ & $\mathcal{L}_{\text{rp}}$ & $\mathcal{L}_{\text{rel}}$ & $\mathcal{L}_{\text{desc}}$ & $\mathcal{L}_{\text{epi}}$
& \textbf{Rep AUC} & \textbf{MS AUC} & \textbf{mAA@10$\degree$} \\
\midrule
 &  &  & $\checkmark$ & $\checkmark$ &  & 64.42 & 61.87 & 28.34 \\
 & $\checkmark$ &  & $\checkmark$ & $\checkmark$ &  & 69.17 & 64.51 & 52.18 \\
 &  & $\checkmark$ & $\checkmark$ & $\checkmark$ &  & 67.25 & 65.15 & 42.90 \\
 &  &  & $\checkmark$ & $\checkmark$ & $\checkmark$ & 67.25 & 62.78 & 53.30 \\
 & $\checkmark$ & $\checkmark$ &  & $\checkmark$ &  & 71.53 & 68.51 & 44.44 \\
 & $\checkmark$ & $\checkmark$ & $\checkmark$ & $\checkmark$ &  & 72.47 & 69.25 & 51.50 \\
 & $\checkmark$ &  & $\checkmark$ & $\checkmark$ & $\checkmark$ & 69.20 & 63.27 & \underline{65.96} \\
 &  & $\checkmark$ & $\checkmark$ & $\checkmark$ & $\checkmark$ & 70.22 & 65.79 & 58.56 \\
\midrule
\textbf{HyKey-noPE} & $\checkmark$ & $\checkmark$ & $\checkmark$ & $\checkmark$ &  & \textbf{82.46} & \textbf{80.68} & 57.48 \\
\textbf{HyKey}      & $\checkmark$ & $\checkmark$ & $\checkmark$ & $\checkmark$ & $\checkmark$ & \underline{75.83} & \underline{72.05} & \textbf{67.18} \\
\bottomrule
\end{tabular*}
\end{table}

Table~\ref{tab:loss-ablation} shows two main effects. First, using only the core matching terms $\mathcal{L}_{\text{rel}}{+}\mathcal{L}_{\text{desc}}$ yields the lowest performance, indicating that descriptor agreement and score reliability alone are insufficient for stable localisation and matching. Adding localisation supervision improves planar performance, with further gains when $\mathcal{L}_{\text{pk}}$ and $\mathcal{L}_{\text{rp}}$ are combined.

Second, $\mathcal{L}_{\text{epi}}$ primarily improves pose accuracy on real video pairs. Across the ablations, epipolar supervision consistently improves relative pose accuracy. This trend is reflected in the full HyKey model, where adding $\mathcal{L}_{\text{epi}}$ improves pose accuracy over HyKey-noPE. Conversely, HyKey-noPE attains the best planar AUC, indicating a trade-off between planar alignment and improved downstream 3D motion estimation.

\subsection{Synthetic Homography Evaluation}\label{subsec:eval}

\begin{table}[t]
\centering
\caption{Homography evaluation on \textit{in-vivo} and \textit{ex-vivo} test sets. We report Rep/MS/MMA/MHA as AUC (best \textbf{bold}, second \underline{underlined}). HyKey variants use HSI; RGB baselines use registered RGB; HS-ALIKE/HSSPN use HSI. ``noPE''~=~ablation without positional/epipolar term. Bold~=~ours.}
\label{tab:syntheticeval}

\begin{tabular*}{\linewidth}{@{\extracolsep{\fill}}llcccc@{}}
\toprule
~ & \textbf{Modality} & \textbf{Rep AUC} & \textbf{MS AUC} & \textbf{MMA AUC} & \textbf{MHA AUC} \\
\midrule
\multicolumn{6}{c}{\textbf{\textit{In-vivo} ColoRectal Dataset}} \\
\midrule
SIFT~\cite{lowe2004distinctive} & RGB & 62.26 & 51.22 & 80.42 & 98.05 \\
SuperPoint~\cite{detone2018superpoint} & RGB & 58.80 & 25.60 & 62.03 & 47.03 \\
RGB-ALIKE~\cite{zhao2022alike} & RGB & 68.05 & 61.35 & 95.03 & 97.42 \\
\midrule
HS-ALIKE & HSI & 64.56 & 57.21 & 93.66 & \underline{98.52} \\
HSSPN~\cite{ma2022deep} & HSI & 66.26 & 45.54 & 76.90 & 79.29 \\
\textbf{HyKey-noPE} & HSI & \textbf{79.17} & \textbf{77.67} & \textbf{97.96} & \textbf{99.45} \\
\textbf{HyKey} & HSI & \underline{71.15} & \underline{67.73} & \underline{95.42} & 98.18 \\
\midrule
\multicolumn{6}{c}{\textbf{\textit{Ex-vivo} Robotic Dataset}} \\
\midrule
SIFT~\cite{lowe2004distinctive} & RGB & 65.77 & 55.66 & 83.25 & 98.62 \\
SuperPoint~\cite{detone2018superpoint} & RGB & 63.28 & 32.67 & 69.87 & 61.86 \\
RGB-ALIKE~\cite{zhao2022alike} & RGB & 70.93 & 64.05 & 95.11 & 98.37 \\
\midrule
HS-ALIKE & HSI & 74.08 & 67.21 & \underline{96.78} & \underline{99.10} \\
HSSPN~\cite{ma2022deep} & HSI & 71.94 & 53.19 & 83.56 & 83.83 \\
\textbf{HyKey-noPE} & HSI & \textbf{82.46} & \textbf{80.68} & \textbf{98.31} & \textbf{99.58} \\
\textbf{HyKey} & HSI & \underline{75.83} & \underline{72.05} & 96.62 & 98.52 \\
\bottomrule
\end{tabular*}
\end{table}

Table~\ref{tab:syntheticeval} summarises homography metrics on the \textit{in-vivo} and \textit{ex-vivo} datasets. HyKey-noPE achieves the highest scores across the homography metrics, while HyKey remains competitive but slightly lower. On the \textit{in-vivo} set, HyKey-noPE outperforms both RGB and HSI baselines across all homography metrics. A similar trend is observed on the \textit{ex-vivo} set, where HyKey-noPE again achieves the best performance and HyKey remains competitive. Both HyKey variants outperform the RGB baselines across all evaluated metrics. These gains reflect improved match density (\ac{ms}) and higher geometric precision (\ac{mma}) in surgical scenes where RGB texture is weak.

This evaluation uses synthetic homography augmentation with known planar warps and photometric variation from differing lighting conditions to assess feature localisation and descriptor accuracy under geometric and illumination changes. Modelling joint spectral-spatial cues with 3D-2D convolutions yields stable descriptors under these transformations, while the slightly higher performance of the noPE variant suggests that epipolar regularisation is less beneficial for purely planar motion. Both HyKey variants demonstrate that hyperspectral features remain discriminative even in visually homogeneous tissue, where \ac{rgb}-based keypoints often fail.

\begin{figure}[t]
    \centering
    \includegraphics[width=0.9\linewidth]{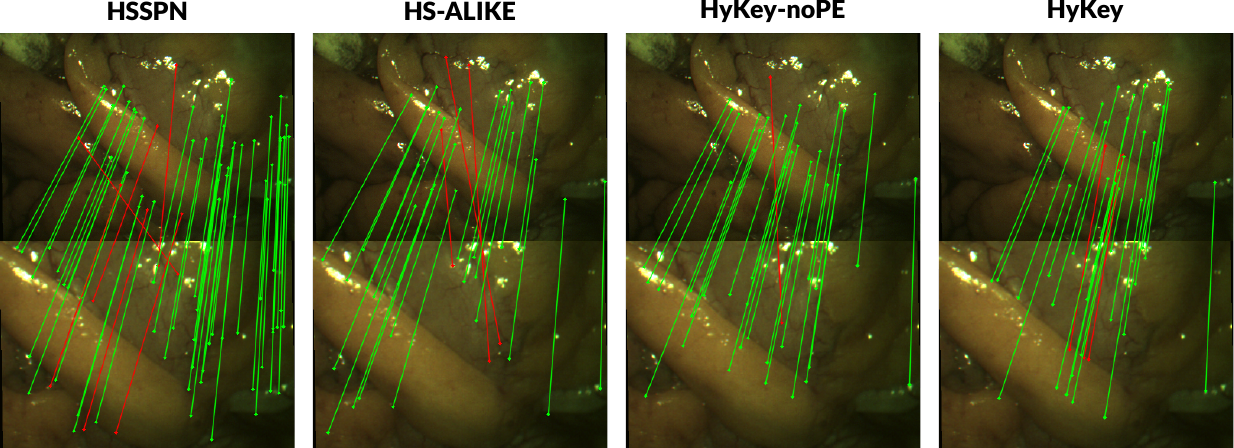}
    \caption{HSI model performance on synthetic homography warping on the \textit{in-vivo} Colorectal dataset. The HyKey models have fewer errors, despite the lack of texture in the scene. Green matches dictate a correct match under three pixels. The number of good and bad matches shown has been proportionally downsampled for visualisation.}
    \label{fig:planar-res}
\end{figure}

Figure~\ref{fig:planar-res} illustrates the difficulty of these planar scenes, where specular glare and limited texture make correspondences ambiguous even to the human eye. HSSPN and HS-ALIKE frequently misalign in specular regions, whereas both HyKey variants maintain stable and accurate matches across these challenging areas.

The HyKey model runs with an average inference \ac{fps} of 23.53~Hz, showing the real-time capabilities of the model. With future optimisation this could be incorporated into a real-time \ac{sfm} pipeline.

\subsection{Real Video Pairs: Relative Pose Evaluation}
\label{subsec:rel-pose-eval}

\begin{table}[t]
\centering
\caption{Relative pose accuracy (mAA) on video sequence pairs using essential matrix recovery. We report mAA@5, mAA@10, mAA@20 (best \textbf{bold}, second \underline{underlined}). HyKey variants and HS-ALIKE/HSSPN use HSI; RGB baselines use registered RGB. ``noPE''~=~ablation without positional/epipolar term. Bold~=~ours.}
\label{tab:nonplanareval}

\begin{tabular*}{\linewidth}{@{\extracolsep{\fill}}lcccc@{}}
\toprule
\textbf{Model} & \textbf{Modality} & \textbf{mAA@5 (\%)} & \textbf{mAA@10 (\%)} & \textbf{mAA@20 (\%)} \\
\midrule
SIFT~\cite{lowe2004distinctive} & RGB & 21.82 & 35.70 & 45.90 \\
SuperPoint~\cite{detone2018superpoint} & RGB & 22.60 & 49.98 & 76.68 \\
RGB-ALIKE~\cite{zhao2022alike} & RGB & 32.56 & 61.56 & \underline{87.70} \\
\midrule
HS-ALIKE & HSI & \underline{37.76} & \underline{63.14} & 87.14 \\
HSSPN~\cite{ma2022deep} & HSI & 32.84 & 59.20 & 83.32 \\
\textbf{HyKey-noPE} & HSI & 32.34 & 57.48 & 79.10 \\
\textbf{HyKey} & HSI & \textbf{38.44} & \textbf{67.18} & \textbf{88.74} \\
\bottomrule
\end{tabular*}
\end{table}

Table~\ref{tab:nonplanareval} reports model performance on real, non-planar video pairs via relative pose estimation. Unlike the synthetic homography tests, where ground-truth point correspondences are known from the applied warp, true sequences do not provide GT correspondences. However, our robotic platform provides ground-truth camera poses, so we assess correspondence quality by relative pose accuracy (mAA@{5$\degree$,10$\degree$,20$\degree$}).

\begin{figure}[t]
    \centering
    \includegraphics[width=0.85\linewidth]{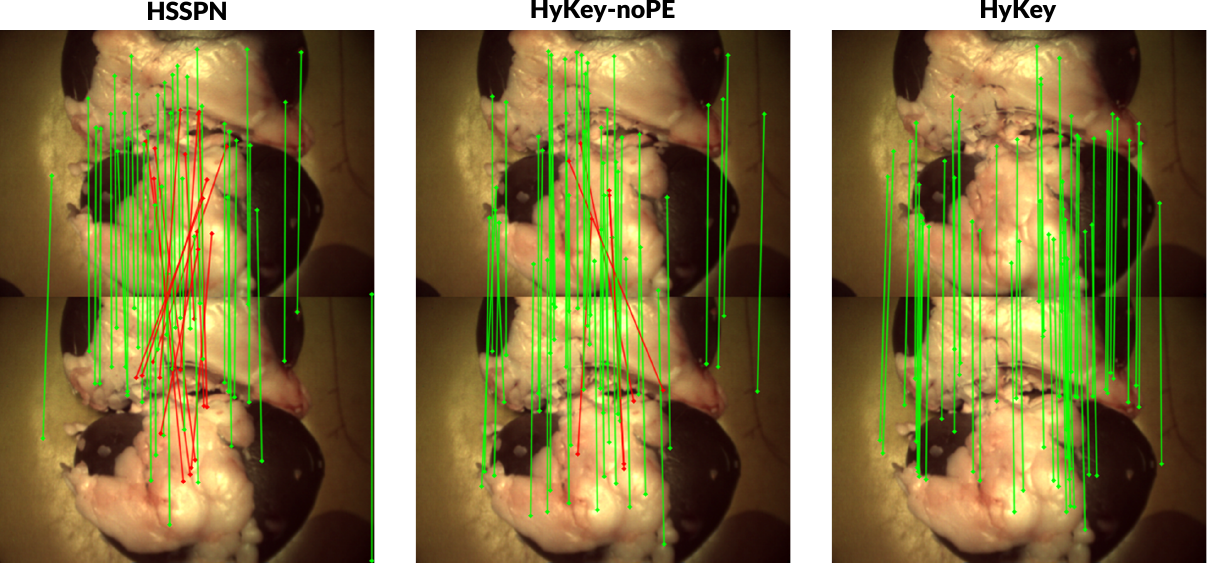}
    \caption{Qualitative view of our model performance on a kidney scene from the robotic \textit{ex-vivo} test set against HSSPN~\cite{ma2022deep}, the only learning-based model for keypoint detection on a pair of video frames. A green match indicates the match has an epipolar error less than 5 pixels. Number of key points downsampled for visualisation.}
    \label{fig:non-planar-res}
\end{figure}

HyKey achieves mAA@10$\degree$~=~67.18\% and mAA@20$\degree$~=~88.74\%, improving over HS-ALIKE (63.14\%, 87.14\%) and SuperPoint (49.98\%, 76.68\%). HyKey-noPE is weaker on pose metrics (57.48\%, 79.10\%), consistent with the intuition that explicit epipolar supervision helps on real non-planar motion, even if pure homography metrics slightly favour our noPE variant. Figure~\ref{fig:non-planar-res} shows the matching performance on the robotic-assisted kidney scene in the test set, where both versions of HyKey outperform HSSPN with fewer erroneous matches.

The largest improvement of HyKey over its no-PE variant appears in relative pose accuracy (mAA), confirming that epipolar supervision strengthens geometric consistency beyond planar alignment. 
Compared to \ac{rgb} models, hyperspectral methods perform best at smaller angular thresholds (mAA@10°), suggesting that the model learns to match tissue regions that are spatially similar but differ subtly in their spectral fingerprints-reflectance patterns invisible to \ac{rgb}. 
Such spectral sensitivity provides a clear advantage for fine-grained correspondence in anatomically homogeneous areas that appear visually uniform in \ac{rgb}. 

Figure~\ref{fig:non-planar-res} shows qualitative results on real laparoscopic pairs from the relative-pose evaluation. HSSPN and HyKey-noPE produce locally correct but directionally inconsistent matches, while the full HyKey variant preserves coherent correspondence directions consistent with the epipolar geometry.

\subsection{Hyperspectral performance against RGB}
\label{subsec:hsi-vs-rgb}

To compare the effect of imaging modality, we compare hyperspectral models (HS-ALIKE, HSSPN) against their closest RGB counterparts (RGB-ALIKE, SuperPoint). Across both synthetic homography and real-pair pose evaluation, hyperspectral methods generally provide stronger geometric correspondences than RGB baselines in these texture-poor, specular conditions typical of surgical scenes.

On the \textit{ex-vivo} robotic dataset (Table~\ref{tab:syntheticeval}), HS-ALIKE improves over RGB-ALIKE on planar localisation and matching, increasing Rep AUC (74.08 vs 70.93) and MS AUC (67.21 vs 64.05) while remaining competitive on MMA/MHA. Similarly, on real non-planar video pairs (Table~\ref{tab:nonplanareval}), HS-ALIKE attains higher mAA@10$\degree$ than RGB-ALIKE (63.14 vs 61.56), indicating improved downstream pose recovery from the recovered correspondences. Although HSSPN does not consistently outperform its RGB analogue in every metric, it remains competitive (for example, mAA@10$\degree$ 59.20 vs SuperPoint 49.98), suggesting that even a 2D CNN trained directly on HSI can benefit from spectral cues.

These trends are consistent with the underlying characteristics of the modality. Compared to RGB, HSI provides additional wavelength information that can help differentiate between visually homogeneous tissue regions and reduce descriptor ambiguity when spatial texture is limited. As a result, hyperspectral keypoints and descriptors can remain more distinctive under challenging \ac{mis} imagery.

\section{Conclusion}\label{sec6}

We presented HyKey, a novel \ac{hsi}-based framework for keypoint detection and description in \ac{mis}, utilising hybrid 3D–2D convolutions for joint spatial–spectral feature learning. 
Through geometry-aware training and benchmarking against \ac{rgb} baselines, HyKey demonstrates that spectral information enhances feature localisation and matching robustness in challenging \ac{mis} scenes. 
Experiments on the proposed dual-modality dataset show consistent improvements in keypoint repeatability, matching precision, and pose recovery, highlighting the potential of \ac{hsi} for reliable geometric perception in surgical vision.
Future work will extend HyKey towards dense reconstruction pipelines using \ac{hsi}-based \ac{sfm} in real dynamic \ac{mis} scenes, as well as towards expanding to other HSI medical systems with different spectral ranges and laparoscopes.

\backmatter

% If your article has accompanying supplementary file/s please state so here. 

% Authors reporting data from electrophoretic gels and blots should supply the full unprocessed scans for key as part of their Supplementary information. This may be requested by the editorial team/s if it is \ac{mis}sing.

% Please refer to Journal-level guidance for any specific requirements.

% \bmhead{Acknowledgements}
% This work was supported in part by the Wellcome/EPSRC Centre for Interventional and Surgical Sciences (WEISS) [203145/Z/16/Z], the Department of Science, Innovation and Technology (DSIT) and the Royal Academy of Engineering under the Chair in Emerging Technologies programme; EPSRC Centre for Doctoral Training in Intelligent, Integrated Imaging In Healthcare (i4health) [EP/S021930/1]; EPSRC Optical and Acoustic imaging for Surgical and Interventional Sciences (OASIS) [UKRI145]. For the purpose of open access, the author has applied a CC BY public copyright licence to any author accepted manuscript version arising from this submission.

\section*{Declarations}

\bmhead{Funding}
This work was supported in part by the Wellcome/EPSRC Centre for Interventional and Surgical Sciences (WEISS) [203145/Z/16/Z], the Department of Science, Innovation and Technology (DSIT) and the Royal Academy of Engineering under the Chair in Emerging Technologies programme; EPSRC Centre for Doctoral Training in Intelligent, Integrated Imaging In Healthcare (i4health) [EP/S021930/1, EP/W00805X/1, EP/Y01958X/1]; EPSRC Optical and Acoustic imaging for Surgical and Interventional Sciences (OASIS) [UKRI145]. 
% \bmhead{Conflict of interest} The authors declare that they have no conflict of interest.

% \bmhead{Ethical approval} This article does not contain any studies with human participants performed by any of the authors.

% \bmhead{Informed consent} This article does not contain patient data.

% Some journals require declarations to be submitted in a standardised format. Please check the Instructions for Authors of the journal to which you are submitting to see if you need to complete this section. If yes, your manuscript must contain the following sections under the heading `Declarations':
% \begin{itemize}
% \item Conflict of interest/Competing interests (check journal-specific guidelines for which heading to use)
% \item Ethics approval and consent to participate
% \item Consent for publication
% \item Materials availability
% \end{itemize}
% }

\noindent

%%===========================================================================================%%

\bibliography{sn-bibliography}% common bib file

@inproceedings{rublee2011orb,
  title={ORB: An efficient alternative to SIFT or SURF},
  author={Rublee, Ethan and Rabaud, Vincent and Konolige, Kurt and Bradski, Gary},
  booktitle={2011 International conference on computer vision},
  pages={2564--2571},
  year={2011},
  organization={Ieee}
}

@inproceedings{detone2018superpoint,
  title={Superpoint: Self-supervised interest point detection and description},
  author={DeTone, Daniel and Malisiewicz, Tomasz and Rabinovich, Andrew},
  booktitle={Proceedings of the IEEE conference on computer vision and pattern recognition workshops},
  pages={224--236},
  year={2018}
}

@article{saikia2025robotic,
  title={Robotic Arm Platform for Multi-View Image Acquisition and 3D Reconstruction in Minimally Invasive Surgery},
  author={Saikia, Alexander and Di Vece, Chiara and Bonilla, Sierra and He, Chloe and Magbagbeola, Morenike and Mennillo, Laurent and Czempiel, Tobias and Bano, Sophia and Stoyanov, Danail},
  journal={IEEE Robotics and Automation Letters},
  year={2025},
  publisher={IEEE}
}

@article{ma2022deep,
  title={A deep learning-based hyperspectral keypoint representation method and its application for 3D reconstruction},
  author={Ma, Tengfei and Xing, Yuxin and Gong, Dawei and Lin, Zijian and Li, Yuanpeng and Jiang, Jiong and He, Sailing},
  journal={IEEE Access},
  volume={10},
  pages={85266--85277},
  year={2022},
  publisher={IEEE}
}

@article{lu2014medical,
  title={Medical hyperspectral imaging: a review},
  author={Lu, Guolan and Fei, Baowei},
  journal={Journal of biomedical optics},
  volume={19},
  number={1},
  year={2014},
  publisher={Society of Photo-Optical Instrumentation Engineers}
}

@article{clancy2020surgical,
  title={Surgical spectral imaging},
  author={Clancy, Neil T and Jones, Geoffrey and Maier-Hein, Lena and Elson, Daniel S and Stoyanov, Danail},
  journal={Medical image analysis},
  volume={63},
  pages={101699},
  year={2020},
  publisher={Elsevier}
}

@inproceedings{balntas2017hpatches,
  title={HPatches: A benchmark and evaluation of handcrafted and learned local descriptors},
  author={Balntas, Vassileios and Lenc, Karel and Vedaldi, Andrea and Mikolajczyk, Krystian},
  booktitle={Proceedings of the IEEE conference on computer vision and pattern recognition},
  pages={5173--5182},
  year={2017}
}

@article{bay2008speeded,
  title={Speeded-up robust features (SURF)},
  author={Bay, Herbert and Ess, Andreas and Tuytelaars, Tinne and Van Gool, Luc},
  journal={Computer vision and image understanding},
  volume={110},
  number={3},
  pages={346--359},
  year={2008},
  publisher={Elsevier}
}

@article{tomasi1992shape,
  title={Shape and motion from image streams under orthography: a factorization method},
  author={Tomasi, Carlo and Kanade, Takeo},
  journal={International journal of computer vision},
  volume={9},
  number={2},
  pages={137--154},
  year={1992},
  publisher={Springer}
}

@article{ullman1979interpretation,
  title={The interpretation of structure from motion},
  author={Ullman, Shimon},
  journal={Proceedings of the Royal Society of London. Series B. Biological Sciences},
  volume={203},
  number={1153},
  pages={405--426},
  year={1979},
  publisher={The Royal Society London}
}

@article{durrant2006simultaneous,
  title={Simultaneous localization and mapping: part I},
  author={Durrant-Whyte, Hugh and Bailey, Tim},
  journal={IEEE robotics \& automation magazine},
  volume={13},
  number={2},
  pages={99--110},
  year={2006},
  publisher={IEEE}
}

@article{zhao2022alike,
  title={Alike: Accurate and lightweight keypoint detection and descriptor extraction},
  author={Zhao, Xiaoming and Wu, Xingming and Miao, Jinyu and Chen, Weihai and Chen, Peter CY and Li, Zhengguo},
  journal={IEEE Transactions on Multimedia},
  volume={25},
  pages={3101--3112},
  year={2022},
  publisher={IEEE}
}

@article{zhao2023aliked,
  title={Aliked: A lighter keypoint and descriptor extraction network via deformable transformation},
  author={Zhao, Xiaoming and Wu, Xingming and Chen, Weihai and Chen, Peter CY and Xu, Qingsong and Li, Zhengguo},
  journal={IEEE Transactions on Instrumentation and Measurement},
  volume={72},
  pages={1--16},
  year={2023},
  publisher={IEEE}
}

@article{lowe2004distinctive,
  title={Distinctive image features from scale-invariant keypoints},
  author={Lowe, David G},
  journal={International journal of computer vision},
  volume={60},
  number={2},
  pages={91--110},
  year={2004},
  publisher={Springer}
}

@article{tejasree2024extensive,
  title={An extensive review of hyperspectral image classification and prediction: techniques and challenges},
  author={Tejasree, Ganji and Agilandeeswari, Loganathan},
  journal={Multimedia Tools and Applications},
  volume={83},
  number={34},
  pages={80941--81038},
  year={2024},
  publisher={Springer}
}

@article{kumar2024deep,
  title={Deep learning for hyperspectral image classification: A survey},
  author={Kumar, Vinod and Singh, Ravi Shankar and Rambabu, Medara and Dua, Yaman},
  journal={Computer Science Review},
  volume={53},
  pages={100658},
  year={2024},
  publisher={Elsevier}
}

@article{ali2025surgical,
  title={Surgical hyperspectral imaging: a systematic review},
  author={Ali, Hafsa Moontari and Xiao, Yiming and Kersten-Oertel, Marta},
  journal={Computer Assisted Surgery},
  volume={30},
  number={1},
  pages={2546819},
  year={2025},
  publisher={Taylor \& Francis}
}

@inproceedings{barath2020magsac++,
  title={MAGSAC++, a fast, reliable and accurate robust estimator},
  author={Barath, Daniel and Noskova, Jana and Ivashechkin, Maksym and Matas, Jiri},
  booktitle={Proceedings of the IEEE/CVF conference on computer vision and pattern recognition},
  pages={1304--1312},
  year={2020}
}

@article{roy2019hybridsn,
  title={HybridSN: Exploring 3-D--2-D CNN feature hierarchy for hyperspectral image classification},
  author={Roy, Swalpa Kumar and Krishna, Gopal and Dubey, Shiv Ram and Chaudhuri, Bidyut B},
  journal={IEEE geoscience and remote sensing letters},
  volume={17},
  number={2},
  pages={277--281},
  year={2019},
  publisher={IEEE}
}

@article{noshiri2023comprehensive,
  title={A comprehensive review of 3D convolutional neural network-based classification techniques of diseased and defective crops using non-UAV-based hyperspectral images},
  author={Noshiri, Nooshin and Beck, Michael A and Bidinosti, Christopher P and Henry, Christopher J},
  journal={Smart Agricultural Technology},
  volume={5},
  pages={100316},
  year={2023},
  publisher={Elsevier}
}

\end{document}